\documentclass[letterpaper, 10 pt, conference]{ieeeconf}  

\IEEEoverridecommandlockouts                              

\usepackage{epsfig}
\usepackage{subfigure}
\usepackage{mathptmx}
\usepackage{times}
\usepackage{amsmath}
\usepackage{amssymb}
\usepackage{amsopn}
\usepackage[ruled,vlined]{algorithm2e}
\usepackage{graphicx}
\usepackage{url}
\usepackage{xcolor} 
\usepackage{balance}
\usepackage{todonotes}

\newcommand\qcp{\textsc{$Q$-CP}}
\newcommand\figurename{Fig.}
\newcommand\set[1]{\textsc{#1}}

\DeclareMathOperator*{\argmax}{\arg\max}

\title{\LARGE \bf \qcp: Learning Action Values for Cooperative
  Planning}

\author{Francesco Riccio$^{1}$ \and Roberto Capobianco$^{1}$ \and
  Daniele Nardi$^{1}$
  \thanks{$^{1}$All authors are with the Department of Computer, Control,
    and Management Engineering ``Antonio Ruberti'', Sapienza
    University of Rome, via Ariosto 25, Rome, 00185, Italy {\tt\small
      {lastname}@dis.uniroma1.it}} }

\begin{document}

\maketitle
\thispagestyle{empty}
\pagestyle{empty}


\begin{abstract}
  Research on multi-robot systems has demonstrated promising results
  in manifold applications and domains. Still, efficiently learning an
  effective robot behaviors is very difficult, due to unstructured
  scenarios, high uncertainties, and large state dimensionality
  (e.g. hyper-redundant and groups of robot). To alleviate this
  problem, we present \qcp{} a cooperative model-based reinforcement
  learning algorithm, which exploits action values to both (1) guide
  the exploration of the state space and (2) generate effective
  policies. Specifically, we exploit $Q$-learning to attack the
  curse-of-dimensionality in the iterations of a Monte-Carlo Tree
  Search. We implement and evaluate \qcp{} on different stochastic
  cooperative (general-sum) games: (1) a simple cooperative navigation
  problem among 3 robots, (2) a cooperation scenario between a pair of
  KUKA YouBots performing hand-overs, and (3) a coordination task
  between two mobile robots entering a door. The obtained results show
  the effectiveness of \qcp{} in the chosen applications, where action
  values drive the exploration and reduce the computational demand of
  the planning process while achieving good performance.
\end{abstract}



\section{Introduction}
\label{sec:intro}

Robots have to show robust behaviors to complete cognitive tasks in
different scenarios, such as service robotics and uncontrolled
industrial environments. However, deciding the best action to perform,
is a complex task due to unpredictabilities of the physical world,
uncertainties in the observations, continuous state spaces and rapid
explosions of the state dimensionality. This is especially true in
multi-robot scenarios such as coordinated navigation and cooperative
manipulation, where each robot has to represent the state of the
environment, and estimate other robots' states in order to determine
the most effective action to perform and, to achieve an individual or
collective goal. In these scenarios, the number of robots induces a
large state-space, where generalization and policy generation become
particularly difficult tasks to tackle.  The generalization problem is
typically addressed with function approximators (e.g. neural
networks), but they do not allow for the use of prior knowledge, which
can be inefficient and lead to dangerous situations. To overcome this
issue, decision theoretical planning techniques, such as Monte-Carlo
tree search~\cite{Browne2012}, have been used to embed prior knowledge
in learning problems. Nevertheless, they show difficulties in relating
similar states (i.e. nodes of the search tree)~\cite{Silver2012}.
Here, we focus on the problem of cooperative general-sum stochastic
games~\cite{Panait2005}, where each agent runs its own learning
process. We attack the generalization problem in policy generation by
enhancing the Upper Confidence Tree (UCT) algorithm~\cite{Kocsis2006}
-- a variant of Monte-Carlo Tree Search -- with an external
action-value function approximator, that selects admissible actions
and consequently drives the node-expansion phase during episode
simulation.

In this paper, we introduce \qcp, Q-value based Cooperative Planning,
to iteratively learn cooperative policies in stochastic games
characterized by discrete action spaces and large state spaces. To
model action values, we use regression on a mixture of
Gaussians~\cite{Agostini2010}, that is iteratively refined by
aggregating new training samples~\cite{Ross2011}, after each search
completion (i.e., at every $t$). Specifically, \qcp{} generates robot
action policies by running multiple Monte Carlo tree
searches~\cite{Winands2008} and incrementally collecting new samples
that are used to improve $Q$-value (i.e., action value)
estimates. These are then used to select admissible actions during the
search process itself. In our experiments, we present a set of fully
collaborative games -- where all the robots have identical reward
functions -- and we therefore choose to not model joint
actions~\cite{Bowling2000, Claus1998}, thus avoiding further
explosions of the UCT search tree.

\begin{figure}[t!]
  \centering
  \subfigure[Cooperative Navigation] {
    \includegraphics[width=0.48\columnwidth]{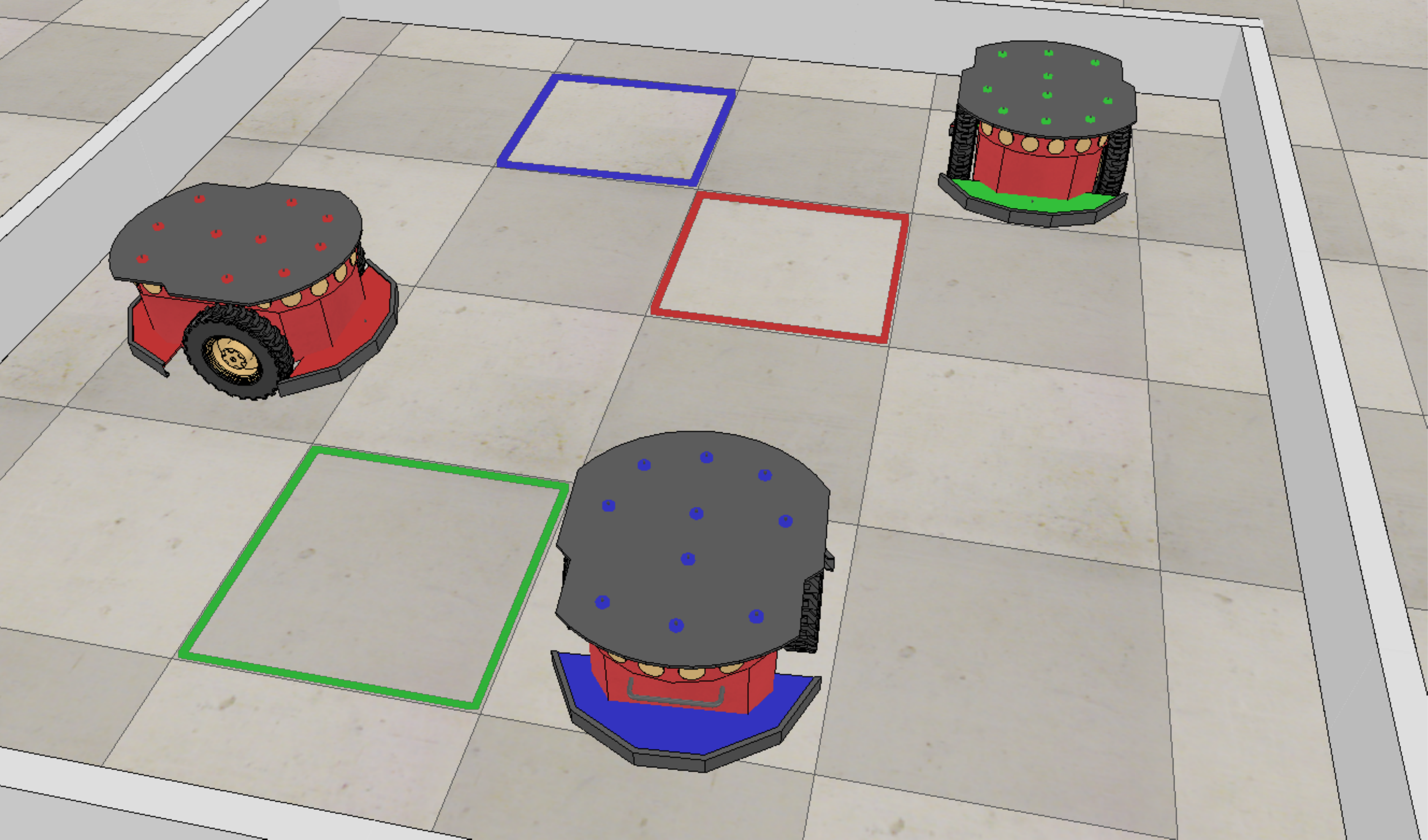}
    \label{fig:simple3_intro}
  }
  \subfigure[Door Passing] {
    \includegraphics[width=0.43\columnwidth]{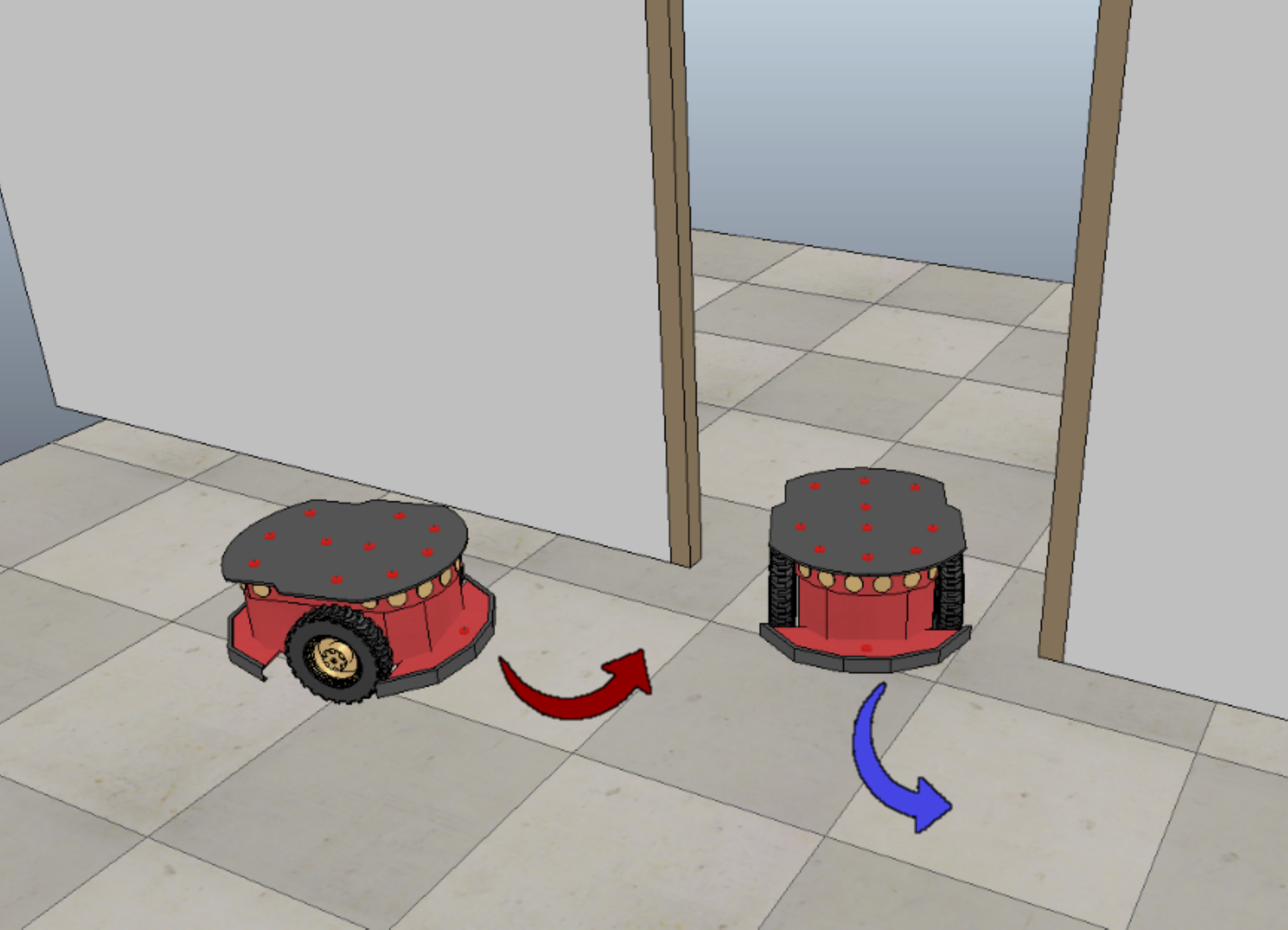}
    \label{fig:door_intro}
  }
  \subfigure[Hand-over] {
    \includegraphics[width=0.8\columnwidth]{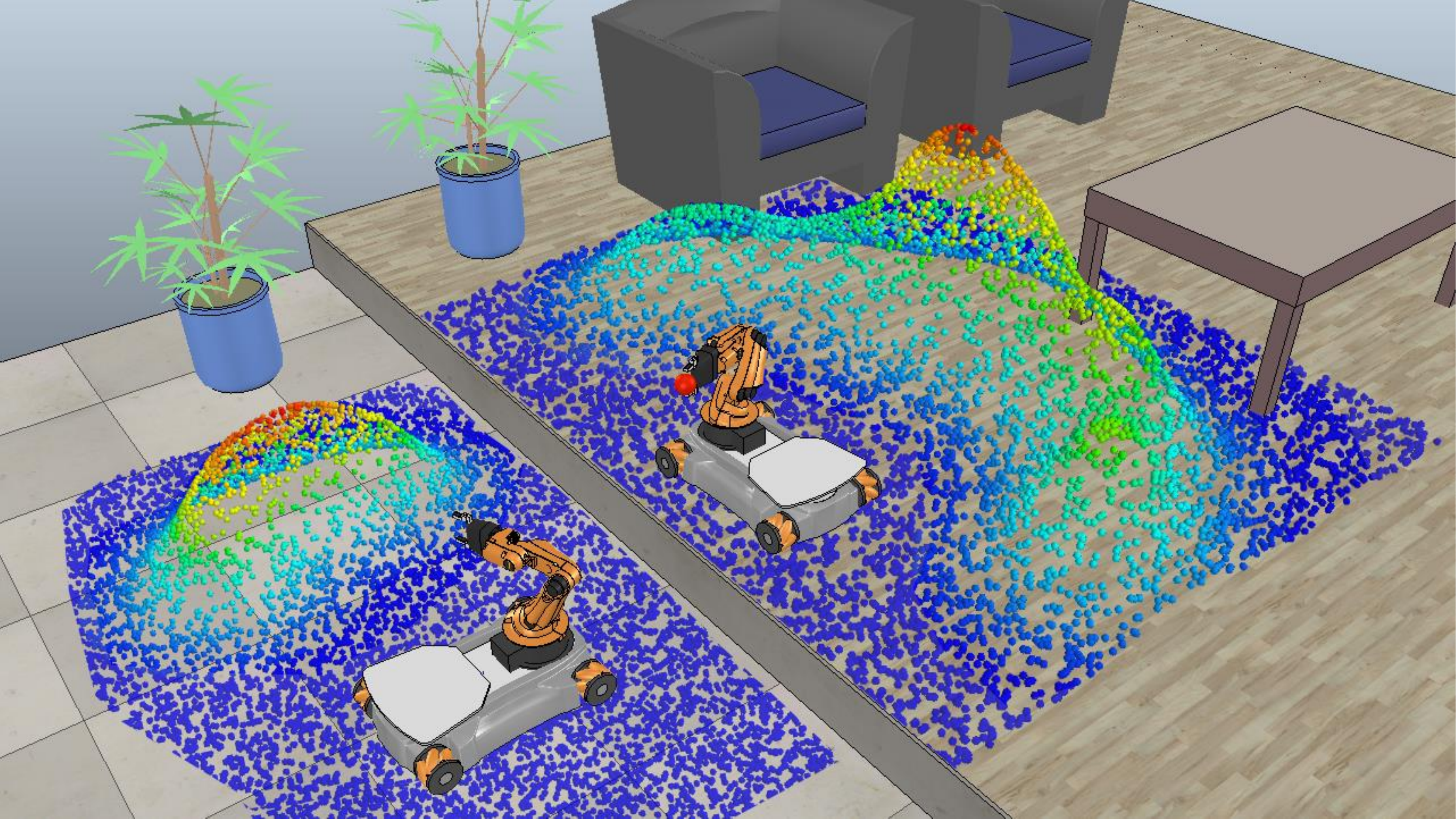}
    \label{fig:coop_youbot_intro}
  }
  \caption{Multi-robot behaviors learned with \qcp. In each of the
    applications, the robots exploit generated policies to succeed in:
    (1) a \textit{cooperative navigation} task
    (\ref{fig:simple3_intro}), where each robot has to reach the
    square with the matching color; (2) \textit{hand-over} task
    (\ref{fig:coop_youbot_intro}) where a robot has to pass a ball
    among each other. The two Gaussian functions model the current
    spatial distribution of two action values generated during
    learning, i.e. \texttt{move-left} for the robot above and
    \texttt{move-right} for the one below; and (3) a \textit{door
      passing} task, where two robots have to coordinate in order to
    traverse a narrow passage in opposite directions.}
  \label{fig:intro}
\end{figure}

We aim at demonstrating that \qcp{} can efficiently be used to
generalize the policy and restrict the search space to support
learning both at an individual and collective level and thus, to
enable teamwork among robots that operate with a common goal. To this
end, we specifically address applications, where cooperation among
robots is necessary to achieve a goal and to improve task performance,
as illustrated in \figurename~\ref{fig:intro}.  The experimental
evaluation shows the effectiveness of explicitly embedding action
values in the search process, resulting in a better generalization and
a reduction in the computational demand of the learning process. Our
main contributions consist in (1) a novel integration of Monte-Carlo
tree search, data aggregation and $Q$-learning, that enables good
performance with large search spaces, (2) an extension of
TD-search~\cite{Silver2012} that constructs upper confidence bounds on
the value function to selects actions optimistically, and (3) a
reduction of the curse-of-dimensionality that is obtained by means of
such a focused exploration. These improvements make our approach more
practical and suitable in difficult robotics applications, where the
lack of training examples is often a limiting condition.

The reminder of the paper is organized as
follows. Section~\ref{sec:related} reports recent research on
reinforcement learning and multi-robot planning, and
Section~\ref{sec:q-cp} describes the \qcp{} algorithm. Finally,
Section~\ref{sec:evaluation} describes our experimental setup, as well
as the obtained results, and Section~\ref{sec:conclusion} closes the
paper with final remarks, limitations and future directions.


\section{Related Work}
\label{sec:related}

Especially in robotics applications where robots are expected to show
robust and adaptable behaviors, generating an effective policy is
often complex and resource intensive, due to large state spaces,
unstructured environments and unpredictabilities. In particular,
techniques that are traditionally adopted in planning or learning are
computationally demanding and time consuming~\cite{Browne2012,
  Kober2013}. To tackle these problems, several approaches initialize
robot policies with reasonable behaviors (e.g., through the aid of
expert demonstrations~\cite{Nikolaidis2015,Kim2016}), or exploit
hierarchical representations like \textit{options}~\cite{Sutton1999,
  Konidaris2016} and MAX-Q
decompositions~\cite{Dietterich2000}. Unfortunately, the applicability
of these methods in the multi-agent domain is rather limited and
unexplored~\cite{Makar2001}. Conversely, in this setting, several
approaches use Q-learning and decentralized POMDPs to solve a
multitude of problems, such as cooperative navigation in 2D grid
worlds~\cite{Abtahi2008,HMenell2016}, the \textit{two agent-tiger} and
a \textit{box pushing} problems~\cite{Bernstein2014}, and \textit{prey
  vs. predator} games in a grid environment~\cite{Proper2009}. These
methods generally require a considerable number of training examples,
and are not easily applicable to robotics domains. Besides being
affected by large state spaces, traditional decision theoretic
planning methods, such as Monte-Carlo tree search, do not provide
generalization capabilities, that are required in common robotics
problems, where large portions of the state space are rarely or never
explored. In robotics, however, generalization may result to be
inefficient, slow and can lead to critical situations due to its
limitation in embedding prior knowledge, which is typically available
to Monte-Carlo techniques~\cite{Browne2012}.

We address generalization at learning time by introducing \qcp, an
iterative algorithm for policy generation. \qcp{} allows a team of
agents to learn action values, that are used for focused
exploration. To this end, we rely on previous
literature~\cite{Engel2005,Chowdhary2014} and we approximate the $Q$
function using probability densities represented as a mixture of
Gaussians~\cite{Agostini2010}. In fact, this type of approximator
provides the flexibility, expressiveness and generality of
non-parametric function approximators, while retaining the additional
benefit of quantifying uncertainty.

Similar to TD-search~\cite{Silver2012}, we aim at reducing the
variance of value estimates during the search procedure by means of
temporal difference learning.  However, we improve our model by
preserving the selective search of Monte-Carlo algorithms, when
bootstrapping is ongoing. Specifically, \qcp{} extends TD-search by
constructing upper confidence bounds on the value function, and by
selecting optimistically with respect to those, instead of performing
$\epsilon$-greedy exploration. In this way, \qcp{} explores more
informative portions of the search space~\cite{Gelly2011, James2017},
and supports the generation of competitive policies -- with the
additional benefit of a reduction in (1) number of simulations (or
expanded nodes), and (2) exploration of the search space -- that
alleviates the curse-of-dimensionality.

Summarizing, our approach enables a team of robots to generate
effective policies directly from simulation, with a reduced training
set and number of iterations. Such a feature makes \qcp{} also
appealing in robotics, where existing approaches to multi-agent
learning cannot be extended nor easily applied.


\section{\qcp}
\label{sec:q-cp}

In this section we present \qcp{} under the stochastic game framework
-- an extension of Markov Decision Processes to the multi-agent
scenario. A stochastic game is a tuple
$(n, \set{S}, \set{A}_{1:n}, \mathcal{T}, R_{1:n})$, where $n$ is the
number of agents, $\set{S}$ is the set of states of the environment,
$\set{A}_{j}$ represents the set of discrete actions of agent $j$,
$\mathcal{T}: \set{S} \times \set{A} \times \set{S} \rightarrow [0,
1]$
is the stochastic transition function that models the probabilities of
transitioning from state $s \in \set{S}$ to $s' \in \set{S}$ when an
action is taken from the joint action space
$\set{A}: \set{A}_{1} \times \dots \times \set{A}_n$, and
$R_{j}: \set{S} \times \set{A} \rightarrow \mathbb{R}$ is the reward
function of agent $i$. In this setting, decisions are represented
through agent policies $\pi_{j}$, that define the behavior of each
agent $j$ by mapping states to actions. Given a stochastic game, the
goal of each agent consists in finding a policy $\pi_{j}(s)$ that
maximizes its future reward with a discount factor $\gamma$.

Our goal is to enable teamwork among robots that operate with a common
goal. For this reason, we restrict to fully collaborative
games~\cite{Bowling2000}, where all the agents have identical reward
functions. \qcp{} iteratively evolves by (1) running, for each agent
an UCT search, where admissible actions are selected through $Q$-value
estimates and (2) incrementally collecting new samples that are used
to improve $Q$-value estimates. In this section, we first describe how
we approximate the $Q$ function; then, we present a detailed
explanation of our algorithm.


\subsection{Preliminaries}
\label{sec:function_approximation}

We choose to approximate the $Q$ function using probability densities
in the form of a mixture of $K$ Gaussians (i.e., Gaussian Mixture
Models -- GMMs), with $K$ determined in an adaptive manner. The use of
such a non-parametric function approximator allows, in fact, to obtain
flexible, expressive and general representations, while retaining the
additional benefit of quantifying uncertainty. To this end, in fact,
we integrate the approach in~\cite{Agostini2010} with a data
aggregation~\cite{Ross2011} procedure, where a dataset of samples is
iteratively collected and aggregated. Specifically, at each iteration
$i$ of \qcp{} we collect a dataset $\set{D}^{i}_{j} = \{x\}$ of sample
state-action pairs and $Q$-values $x = (s,a_{j},Q^{i}_{j}(s,a_{j}))$
experienced by agent $j$, together with the corresponding rewards $r$
and transitioned states $s'$ resulting from the executed joint
action. The values $Q^i_{j}$ are determined, at each iteration $i$,
according to the $Q$-learning update rule
\begin{align}
  \small
  \begin{split}
    \label{eq:q_rule_update}
    Q^{i}_{j}(s,a_{j}) =~& \hat{Q}^{i-1}_{j}(s,a_{j}) \\ & + \alpha~(r +
    \gamma\max_{a'_{j}}\hat{Q}^{i-1}_{j}(s',a'_{j}) -
    \hat{Q}^{i-1}_{j}(s,a_{j})),
  \end{split}
\end{align}
where $\alpha$ is the learning rate, $\hat{Q}^{i-1}_{j}$ is the
function approximation, and $Q^{0}_{j}(s,a_{j}) = 0$. As we discuss
later, the function approximation $\hat{Q}$ is learned over an
aggregated dataset
$\set{D}^{0:i}_{j} = \{\cup\set{D}^{d}_{j}|d=0 \dots i\}$. More
specifically, the aggregated dataset is used to estimate a probability
density function (pdf) in the space of states, actions and
$Q$-values.

The pdf depends on the set of parameters
${\Theta} = \{\pi_{1},\mu_{1},\Sigma_{1},\dots,\pi_{K},
\mu_{K},\Sigma_{K}\}$,
where $\pi_{k}$ is the prior, $\mu_{k}$ the mean and $\Sigma_{k}$ the
covariance matrix of a Gaussian of dimensionality $G$ in the canonical
form
\begin{align}
  \small
  \set{N}({x},\mu,\Sigma) = 
  \dfrac{1}{\sqrt{(2\pi)^G|\Sigma|}}e^{-\frac{1}{2}({x}-\mu)^T\Sigma^{-1}({x}-\mu)}.
\end{align}
The parameters ${\Theta}$ are estimated over the dataset
$\set{D}^{0:i}_{j}$, and they are obtained as the result of a standard
Expectation-Maximization~\cite{Dempster1977} procedure, initialized
using the k-means algorithm~\cite{MacQueen1967} to avoid bad local
optima. The number of components $K$ of the GMM is selected to
minimize the Bayesian Information Criterion~\cite{Schwarz1978} over a
testing portion of the dataset.

Using ${\Theta}$, the pdf of a sample can be computed as
\begin{align}
  \small
  p({x};{\Theta}) = \sum^K_{k=1}\pi_k\set{N}
  ({x},\mu_k,\Sigma_k),
\end{align} 
while the value $\hat{Q}(s,a) \approx Q(s,a)$ is obtained as
$\hat{Q}(s,a) = \mathbb{E}[Q|s,a] = \mu(Q|s,a)$. The approximated
$Q$-value is the result of a Gaussian Mixture Regression
\begin{align}
  \small
  \begin{split}
    \hat{Q}(s,a) = \mu(Q|s,a) = \sum^K_{k=1}\beta(s,a)_k\mu_k(Q|s,a), \\
    \sigma^2(Q|s,a) = \sum^G_{g=1}\beta(s,a)_gy_g(s,a)- \mu^2(Q|s,a),
  \end{split}
\end{align}
that originates from the decomposition of each $\mu_{k}$ and
$\Sigma_{k}$ into
\begin{align}
  \small
  \mu_{k} = \Bigg(
  \begin{array}{l}
    \mu^{(s,a)}_{k}\\
    \mu^{Q}_{k}
  \end{array}
  \Bigg)~~
  \Sigma_{k} = \Bigg(&
                       \begin{array}{l l}
                         \Sigma^{(s,a)(s,a)}_{k} & \Sigma^{(s,a),Q}_{k}\\
                         \Sigma^{Q,(s,a)}_{k} & \Sigma^{Q,Q}_{k}
                       \end{array}
                                                \Bigg),
\end{align}
where $\mu_{k}(Q|s,a)$, $\sigma^2_{k}(Q|s,a)$, $y_g(s,a)$ and
$\beta_{k}(s,a)$ are respectively
\begin{align}
  \small
  \mu_{k}(Q|s,a) = &\mu^Q_{k} + \Sigma^{Q,(s,a)}_{k}(\Sigma^{(s,a)}_{k})^{-1}((s,a)-\mu^{(s,a)}_{k})\\
  \sigma^2_{k}(Q|s,a) &= \Sigma^{Q,Q}_{k} -
                        \Sigma^{Q,(s,a)}_{k}(\Sigma^{(s,a)}_{k})^{-1}\Sigma^{(s,a),Q}_{k}\\
  y_g(s,a) &= (\sigma^2_g(Q|s,a) + \mu^2_g(Q|s,a))\\
  \beta_{k}(s,a) &=
                   \dfrac{\set{N}(s,a;\mu^{(s,a)}_{k},\Sigma^{(s,a),(s,a)}_{k})}
                   {\sum^K_{k=1}\set{N}(s,a;\mu^{(s,a)}_{k},\Sigma^{(s,a),(s,a)}_{k})}.
\end{align}


\subsection{Algorithm}
\label{sec:q-cp-algo}

\qcp{} is a planning algorithm that uses Q-values to effectively
explore the state space $S$ in a Monte-Carlo tree search, with the
goal of generating action policies $\pi_j$ for each agent $j$ that
operates in a team with a common goal. At each iteration $i$, \qcp{}(1)
runs -- for each agent -- an UCT search, where admissible actions are
selected through $Q$-value estimates, and (2) incrementally collects
new samples that are used to improve $Q$-value estimates according to
the procedure (as described in previous section).

More in detail (see Algorithm~\ref{alg:q-cp}), \qcp{} takes as input an
initial policy $\pi^0_j$ for each agent $j$ and, at each iteration
$i = 1 \dots I$, evolves as follows:
\begin{enumerate}
\item all the agents simultaneously follow their policy
  $\pi^{i-1}_{j}$ for $T$ timesteps, generating a set of $T$ states
  $\{s_t\}$.

\item for each generated state $s_t$, each agent $j$ sequentially runs
  a modified UCT search with depth $H$. Specifically, at each
  $h = 1 \dots H$, the search algorithm

  \begin{enumerate}
  \item evaluates a subset of ``admissible'' actions
    $\tilde{\set{A}}_{j} \subseteq \set{A}_{j}$ in $s_{t+(h-1)}$.
    Admissible actions are determined according to
    $\hat{Q}(s_{t+(h-1)},a_j)$ and $\Sigma(Q|s_{t+(h-1)},a_j)$. In
    particular, differently from vanilla UCT, we only allow actions
    $a_j$ such that
    \begin{align}
      \begin{split}
        \small \hat{Q}(s_{t+(h-1)},a_j)& >=~
        \lambda\max_{a}\hat{Q}^{i-1}_j(s_{t+(h-1)}, a) - \delta \\ &
        \delta\sim\sigma^2(Q|s_{t+(h-1)},a_j),
      \end{split}
    \end{align}
    \noindent where $\lambda$ is typically initialized to $0.5$ and
    increases with the number of iterations $i$. Through $\delta$, the
    prediction error for each action is captured, leading to a more
    directed exploration strategy. Additionally, an action can be
    randomly determined to be admissible with
    $\epsilon_{\tilde{\set{A}}}$ probability.

  \item selects and executes the best action
    $a^{h*}_j \in \tilde{\set{A}}_{j}$ according to
    \begin{align}
      \small
      \begin{split}
        \label{eq:uct_max_action}
        e &= C \cdot
        \sqrt{\frac{\log(\sum_{a_j}{\eta(s_{t+(h-1)},a_{j})})}{\eta(s_{t+(h-1)},a_{j})}} \\
        a^{h*}_j &= \argmax_{a_j} \hat{Q}^{i-1}_j(s_{t+(h-1)}, a_{j}) + e,
      \end{split}
    \end{align}
    where $C$ is a constant that multiplies and controls the
    exploration term $e$, and $\eta(s_{t+(h-1)},a_{j})$ is the number
    of occurrences of $a_{j}$ in $s_{t+(h-1)}$. To use \qcp{} on
    robotic applications, where the state space $\set{S}$ is typically
    continuous, we define a comparison operator that introduces a
    discretization by informing the algorithm whether the difference
    of two states is smaller than a given threshold $\xi$.

  \item runs $M$ simulations (or roll-outs), by simultaneously
    executing the $\epsilon$-greedy policies of each agent $j$ --
    based on $\pi^{i-1}_{j}$ -- until a terminal state is reached. It
    is crucially important to remark that during each roll-out, all
    the robots execute their policy. In this way, \qcp{} evaluates the
    optimal action according teammates' expected actions.
  \end{enumerate}

  \qcp{} uses UCT as an expert and collects, at each iteration and for
  each agent, a dataset $\set{D}^{i}_{j}$ of $H$ samples
  $x = (s,a_{j},Q^{i}_{j}(s,a_{j}))$.

\item after each iteration of UCT, each agent aggregates
  $\set{D}^{i}_{j}$ into its own
  $\set{D}^{0:i}_{j} = \set{D}^{i}_{j} \cup
  \set{D}^{0:i-1}_{j}$~\cite{Ross2011,Ross2014},
  and uses the complete dataset to learn the $Q$-values, as
  illustrated in previous section. It is worth remarking that selected
  actions influence the distribution of states and make the dataset
  non-i.i.d.

\item once $\hat{Q}^{i}_j$ is learned, the policy is also updated as
  $\pi^{i}_{j}(s) = \argmax_{a} \hat{Q}^{i}_j(s,a)$.
\end{enumerate}

\begin{algorithm}[!t]
  \small
  \DontPrintSemicolon \SetAlgoLined 
  \SetNlSty{}{}{)}
  \SetAlgoNlRelativeSize{0}

  \KwData{$n$ number of agents; $I$ the number of iterations; $\Delta$
    initial state distribution; $H$ UCT horizon; $T$ policy execution
    timesteps, $\lambda_0$ initial max. $Q$ threshold multiplier for
    admissible actions, $\epsilon_{\tilde{\set{A}}}$ probability for
    random admissible actions, $\alpha$ learning rate, $\gamma$
    discount factor.}

  \KwIn{$\pi^0_j$ initial policy for each agent $j$}

  \KwOut{$\pi^{I}_{j}$ policy learned after I iterations}
  
  \Begin{
    \For{$i = 1$ \KwTo $I$} {
      $s_0 \leftarrow$ random state from $\Delta$.\;
      \For{$t = 1$ \KwTo $T$} {
        \nl Get state $s_t$ by executing $\pi^{i-1}_{j}(s_{t-1})$ for
        each agent $j$.\;

        \For{$j = 1 \dots n$} {
          \nl $\set{D}^{i}_{j} \leftarrow$ UCT$_{\qcp}$($s_t,
          \lambda_0, \epsilon_{\tilde{\set{A}}}$).\;
          \nl $\set{D}^{0:i}_{j} \leftarrow \set{D}^{i}_{j} \cup
          \set{D}^{0:i-1}_{j}$.\;
          $\hat{Q}^{i}_j \leftarrow$
          function\_approximation($\set{D}^{0:i}_{j}, \alpha, \gamma$).\;
          \nl $\pi^{i}_{j}(s) \leftarrow \argmax_{a} \hat{Q}^{i}_j(s,a)$.\;
        }
      }
    }

    \BlankLine
    \Return{$\pi^{I}_{j}$}\; 
  }
  \caption{\qcp}
  \label{alg:q-cp}
\end{algorithm}


\section{Experimental Evaluation}
\label{sec:evaluation}

Generating competitive policies in a multi-robot environment is a
challenging task, due to the exploration of large state spaces. Hence,
in order to be practical in multiple robotic applications, the goal of
\qcp{} is to enable a team of robots to plan an effective policy with
a reduced number of explored states and algorithm iterations. Thus
improving the simulation time and computational load. To this end, we
compare our solution against TD-search~\cite{Silver2012} and both a
\textit{vanilla-UCT} and \textit{random-UCT} implementations. We refer
to \textit{vanilla-UCT} as the standard UCT algorithm that, at each
iteration, expands every possible action in $A_j$, for every agent
$j$. The \textit{random-UCT} instead, is a baseline algorithm, where
at each step of the UCT search, it randomly selects only one action to
expand. Specifically, we evaluate \qcp{} in the three scenarios shown
in \figurename~\ref{fig:intro}: (1) a \textit{cooperative navigation}
application, where three robots have to coordinate in order to find
the minimum path to their respective targets in a grid world; (2) a
\textit{hand-over task}, where two robots have to overcome structural
constraints in order to complete their job; and (3) a
\textit{coordination task} among two robots passing through a door in
opposite directions -- this is a challenging setup, since timing and
the selection of the correct action are critical.


\subsection{Experimental Setup}
\label{subsec:setup}

Experiments have been conducted using the V-REP simulator running on a
single Intel Core i7-5700HQ core, with CPU@2.70GHz and 16GB of RAM.
For all the scenarios, unless otherwise specified, the algorithm was
configured with the same meta-parameters. The UCT horizon is set to
$H = 4$ to trade-off between usability and performance of the search
algorithm; the number of roll-outs is set to be $3$, while admissible
actions are evaluated with an initial $\lambda = 0.5$, and
$\epsilon_{\tilde{\set{A}}} = 0.3$, guaranteeing good amounts of
exploration. The $C$ constant in Eq.~\ref{eq:uct_max_action} is set to
$0.7$.  The learning rate $\alpha$ in Eq.~\ref{eq:q_rule_update} is
set to $0.2$, while the discount factor $\gamma$ is equal to
$0.8$. The number of $K$ components for the GMM approximator is 5 for
the first two scenarios and 6 for \textit{door passing}. The state
space, the set of actions and the reward functions are finally
implemented depending on the robots and applications. In each of the
proposed applications, stochastic actions are obtained by randomizing
the their outcomes with a 5\% probability. We evaluate the reward
obtained during different executions of \qcp{} against the number of
explored states and iteration of the algorithms. In particular, the
learning system implements a \textit{shaped} reward function, which
computes a reward value for each of the visited states.

\subsection{Cooperative Navigation}
\label{sec:cooperative_navigation}
\begin{figure}[t!]
  \centering
  \subfigure[Rewards] {
    \includegraphics[width=\columnwidth]{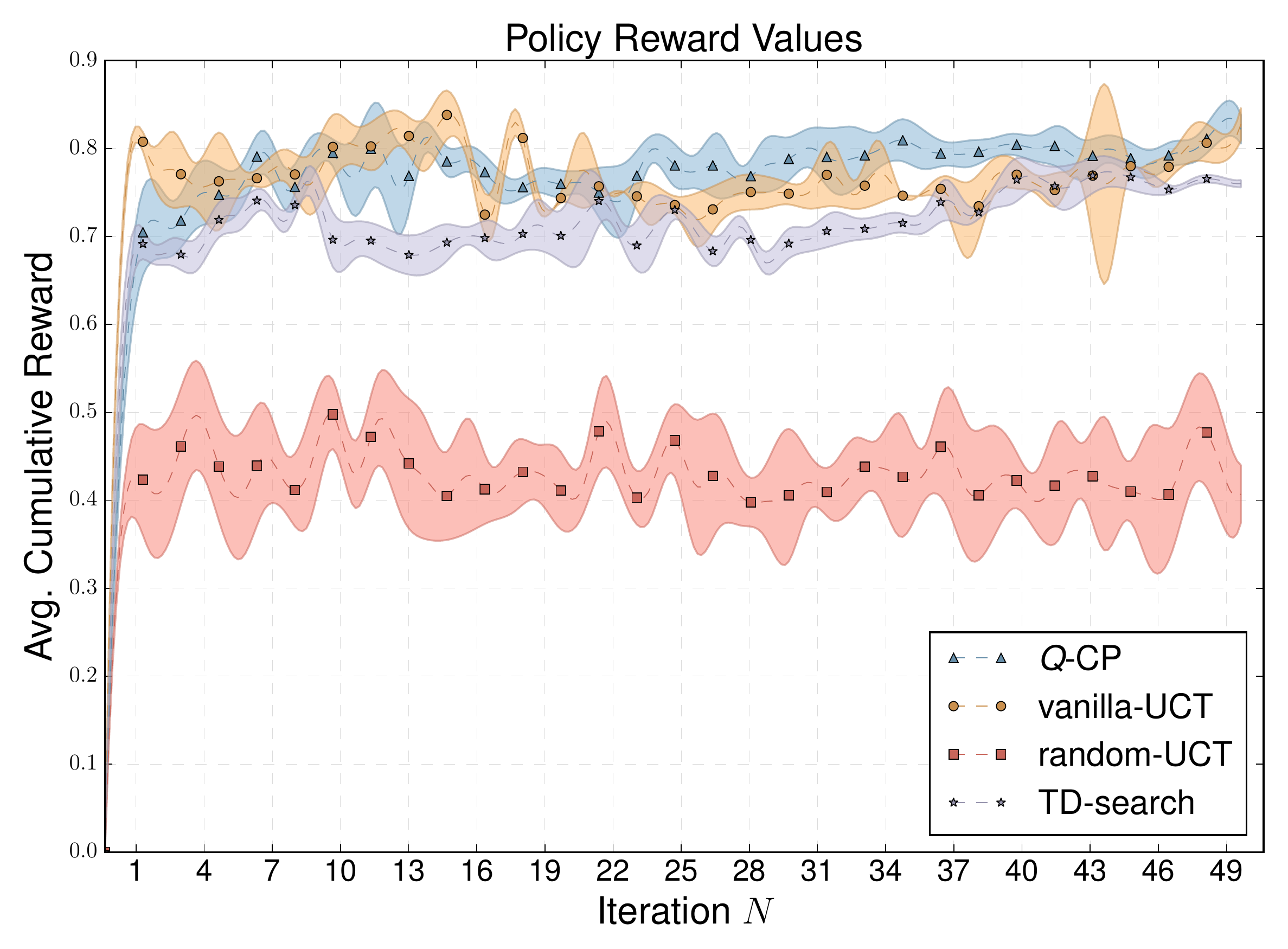}
    \label{fig:simple3_rewards}
  }
  \subfigure[States] {
    \includegraphics[width=\columnwidth]{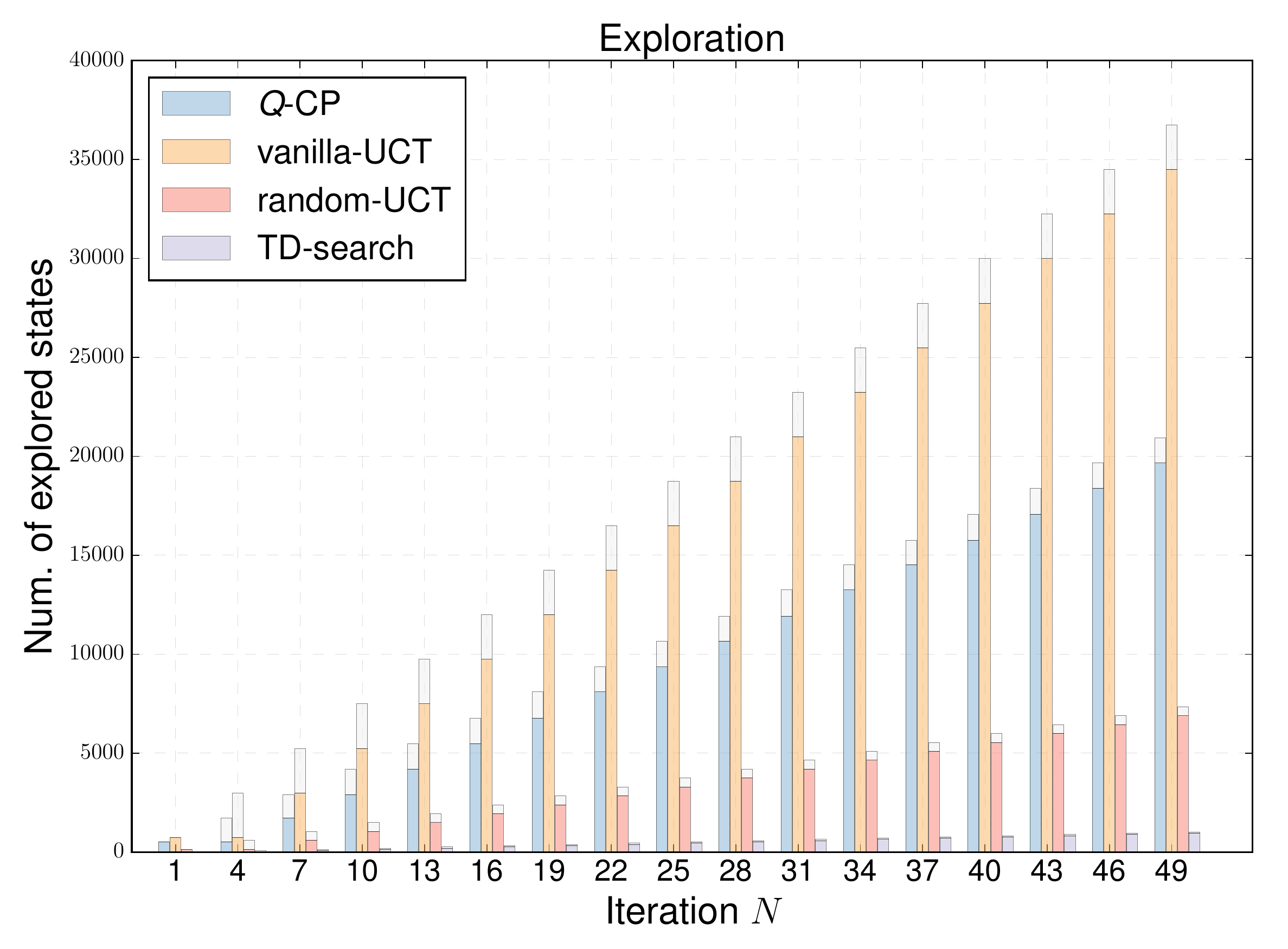}
    \label{fig:simple3_states}
  }
  \caption{Normalized average reward (a) and number of explored states
    (b) obtained by \qcp, TD-search, \textit{random-UCT} and
    \textit{vanilla-UCT} in 49 iterations in the cooperative
    navigation scenario. For each of them, the reward is averaged over
    10 runs. The averaged reward is represented as a dotted line
    while, the line width represent its standard deviation. At each
    iteration, the number of explored states is represented as a bar,
    where the height indicates the total number of visited states, and
    the gray top highlights the amount of states added during the
    particular iteration $i$. The plot shows a comparison among
    \qcp{}, TD-search, \textit{random-UCT} and \textit{vanilla-UCT}.}
  \label{fig:simple3}
\end{figure}

In this scenario, three robots have to perform a cooperative
navigation task in a 4x4 grid world
\figurename~\ref{fig:simple3_intro}. Specifically, their task consists
in reaching the square that matches their color (highlighted in the
figure) by avoiding collisions with teammates. The state is a
12-features vector, where each agent is represented by 4 features
encoding its current and target cells
$s = \big\langle r_x,~ r_y,~ t_x,~ t_y \big\rangle$, while the set of
actions, for each robot, is composed as a tuple of 5 elements
$A = \langle$ \texttt{noop, up, down, right, left} $\rangle$. The
reward function is implemented to be inversely proportional to the sum
of the Manhattan distances among robots and their targets.
\figurename~\ref{fig:simple3} shows the comparison of \qcp{} with
TD-search, \textit{random-UCT},
\textit{vanilla-UCT}. \figurename~\ref{fig:simple3_rewards} shows the
normalized average reward (dotted line) and standard deviation (line
width) of the considered algorithms obtained in 49 iterations and
averaged over 10 runs for each of them.
\figurename~\ref{fig:simple3_states}, shows the number of states
explored per iteration. Each bar in the plot highlights the total
number of states explored until the i\textit{-th} iteration, and the
gray top highlights the amount of states expanded during $i$ with
respect to the total number of states explored until $i-1$. As
expected, \textit{random-UCT} explores a low number of states -- since
it expands only one random action at each UCT iteration -- but it
performs poorly. Conversely, if we consider \qcp{} and
\textit{vanilla-UCT}, we can notice similar reward
values. Nonetheless, our solution generates a competitive policy with
a remarkably reduced number of explored states. Although in this
scenario TD-search explores less states than \qcp{}, obtaining
comparable rewards, the state space is relatively small (i.e. 4x4 grid
world) and the problem can be solved by greedily following the
estimated value function. In fact, once a first approximation is found
at every state, following the max valued action matches the global
optimum. As we will see in the following scenarios, however, TD-search
reports a worse performance as the state dimensionality increases and,
a greedy search over approximated value functions is no longer
sufficient in both of them. In the case of the door passing experiment
\figurename~\ref{fig:door_intro}, in which immediately following the
estimated value function -- without allowing optimistic exploration --
leads to deadlock in the robot policies. For example, if the goal is
on the opposite side of the wall and the door is far from the robot,
the agent needs to visit states that do not necessarily maximize a
first approximation of the value function.  Conversely, visiting these
states (as \qcp{} does by constructing upper confidence bounds of the
value function), allows for a better approximation and thus for an
improved policy.

\subsection{Hand-over}
\begin{figure}[t!]
  \centering
  \subfigure[Rewards] {
    \includegraphics[width=\columnwidth]{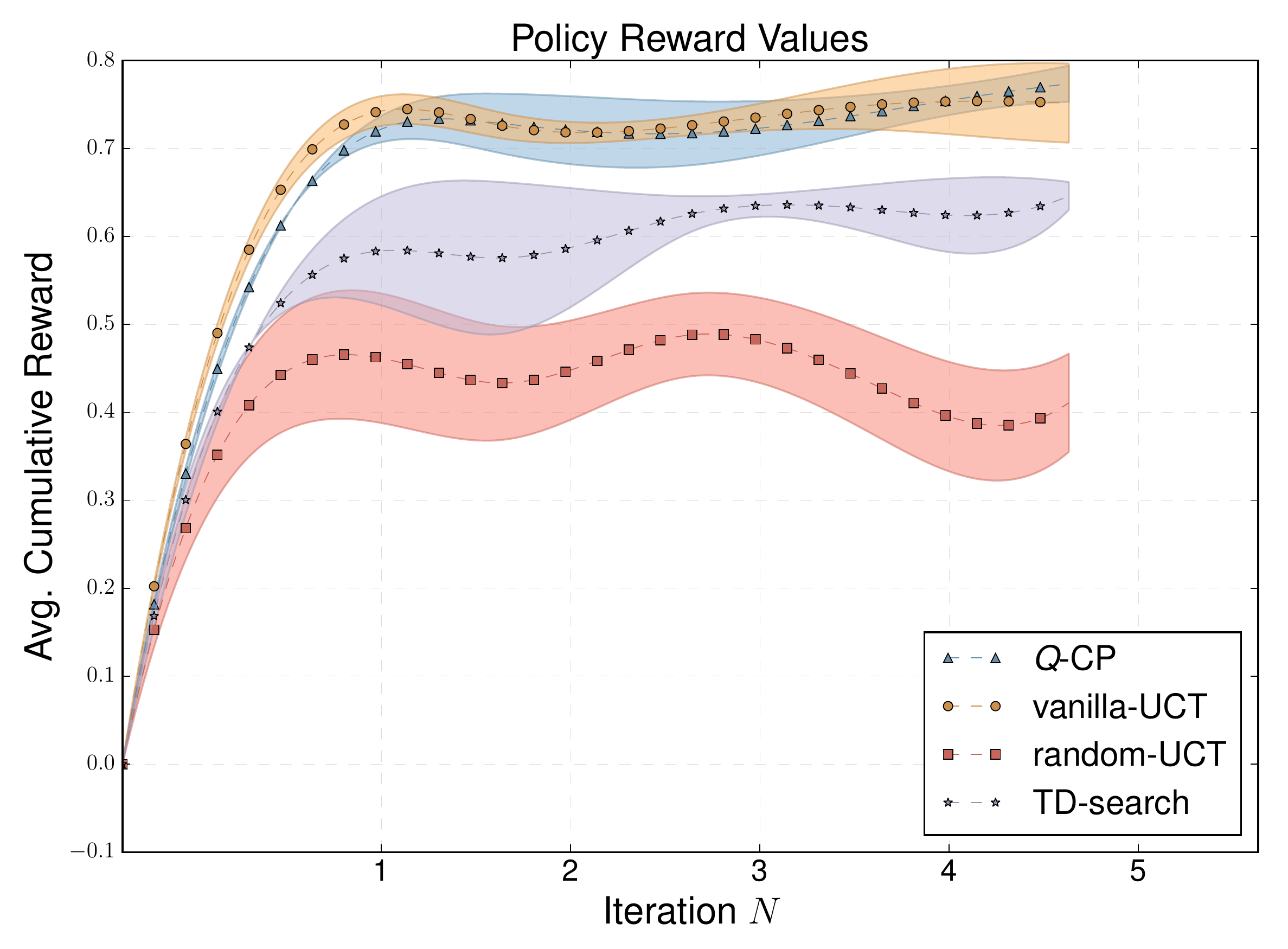}
    \label{fig:coop_youbot_rewards}
  }
  \subfigure[States] {
    \includegraphics[width=\columnwidth]{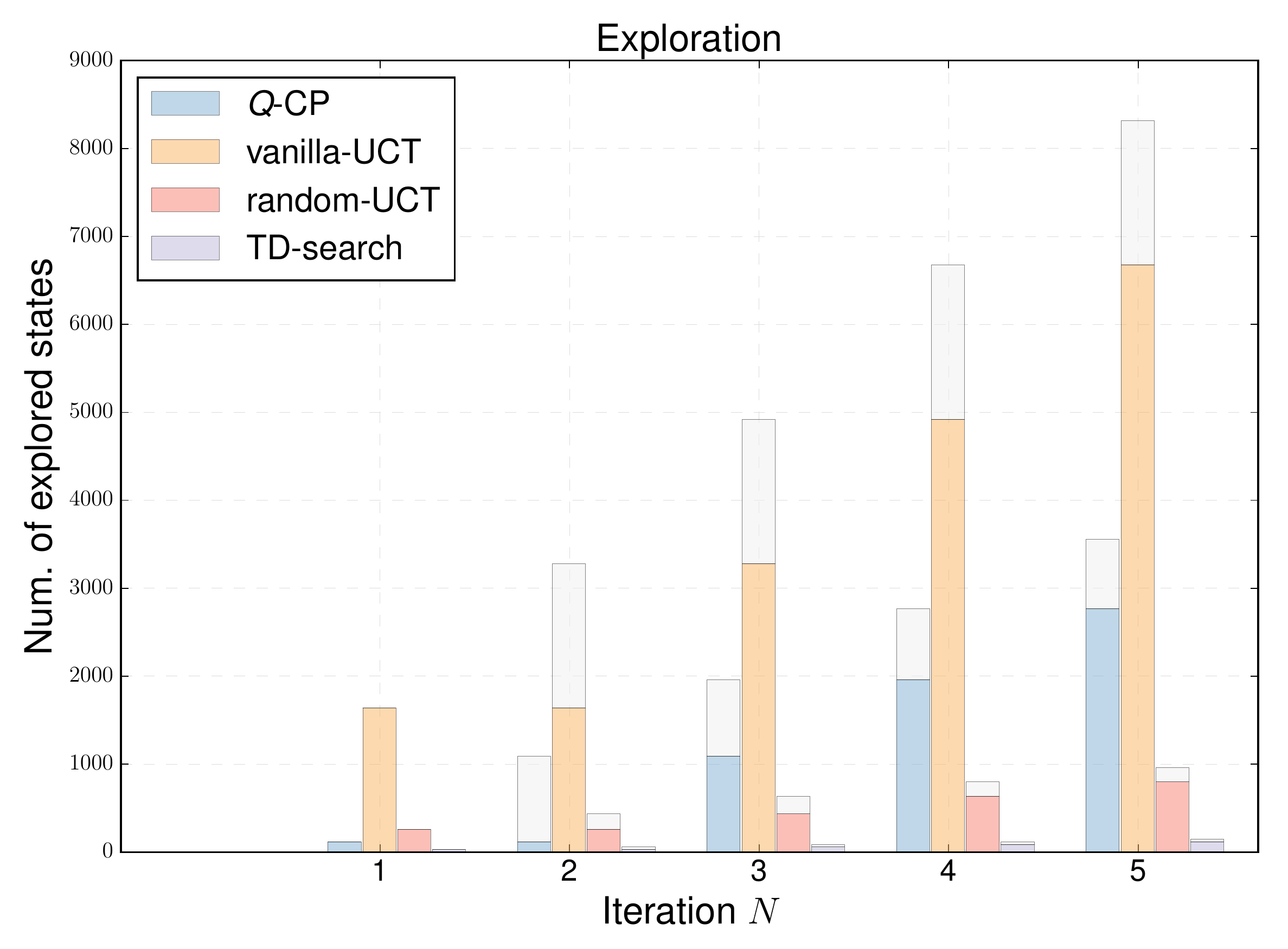}
    \label{fig:coop_youbot_states}
  }
  \caption{Normalized average reward and number of explored states
    obtained by \qcp{}, TD-search, \textit{random-UCT} and
    \textit{vanilla-UCT} in 5 iterations in the hand-over scenario.
    For each of them, the reward is averaged over 10 runs.}
  \label{fig:coop_youbot}
\end{figure}
In this scenario, two robots have to learn how to complete a hand-over
task in order to move an object
(\figurename~\ref{fig:coop_youbot_intro}). Here, the state space is
inherently continuous and is represented as a vector of 5 elements
encoding the relative distance of both the robot bases (2D vector) and
the end-effectors of their 5DOF arms (3D vector),
$s = \langle rel\_base_{x},~ rel\_base_{y},~ rel\_end-effector_{x},~
rel\_end\_effector_{y},~ rel\_end\_effector_{z} \rangle$.
Note that, for running UCT, a discretization is performed as described
in Section~\ref{sec:q-cp-algo}.  The set of 10 actions, for each
robot, consists of $A = \langle$ \texttt{arm-up, arm-down,
  arm-forward, arm-backward, arm-right, arm-left, base-forward,
  base-backward, base-left, base-right} $\rangle$. The reward function
is implemented as a weighted sum of 2 components inversely
proportional to the Euclidean distance between (1) the robot bases and
a desired distance (1.0 m), and (2) the arm end-effectors. Hence, the
reward promotes states where the two bases maintain a distance of one
meter and their end-effectors are as close as possible to enable the
hand-over of the object. As in the previous scenario,
\figurename~\ref{fig:coop_youbot} shows a comparison between \qcp{},
TD-search, \textit{random-UCT} and \textit{vanilla-UCT}. This scenario
further highlights the ability of \qcp{} in generating competitive
policies with a reduced number of explored states. While TD-search and
\textit{random-UCT} report worse performance in terms of obtained
reward, \textit{vanilla-UCT} has more competitive values
(\figurename~\ref{fig:coop_youbot_rewards}). However, the number of
expanded nodes is significantly lower ($<\sim45\%$) since first
iterations.

\subsection{Door Passing}
\begin{figure}[t!]
  \centering
  \subfigure[Rewards] {
    \includegraphics[width=\columnwidth]{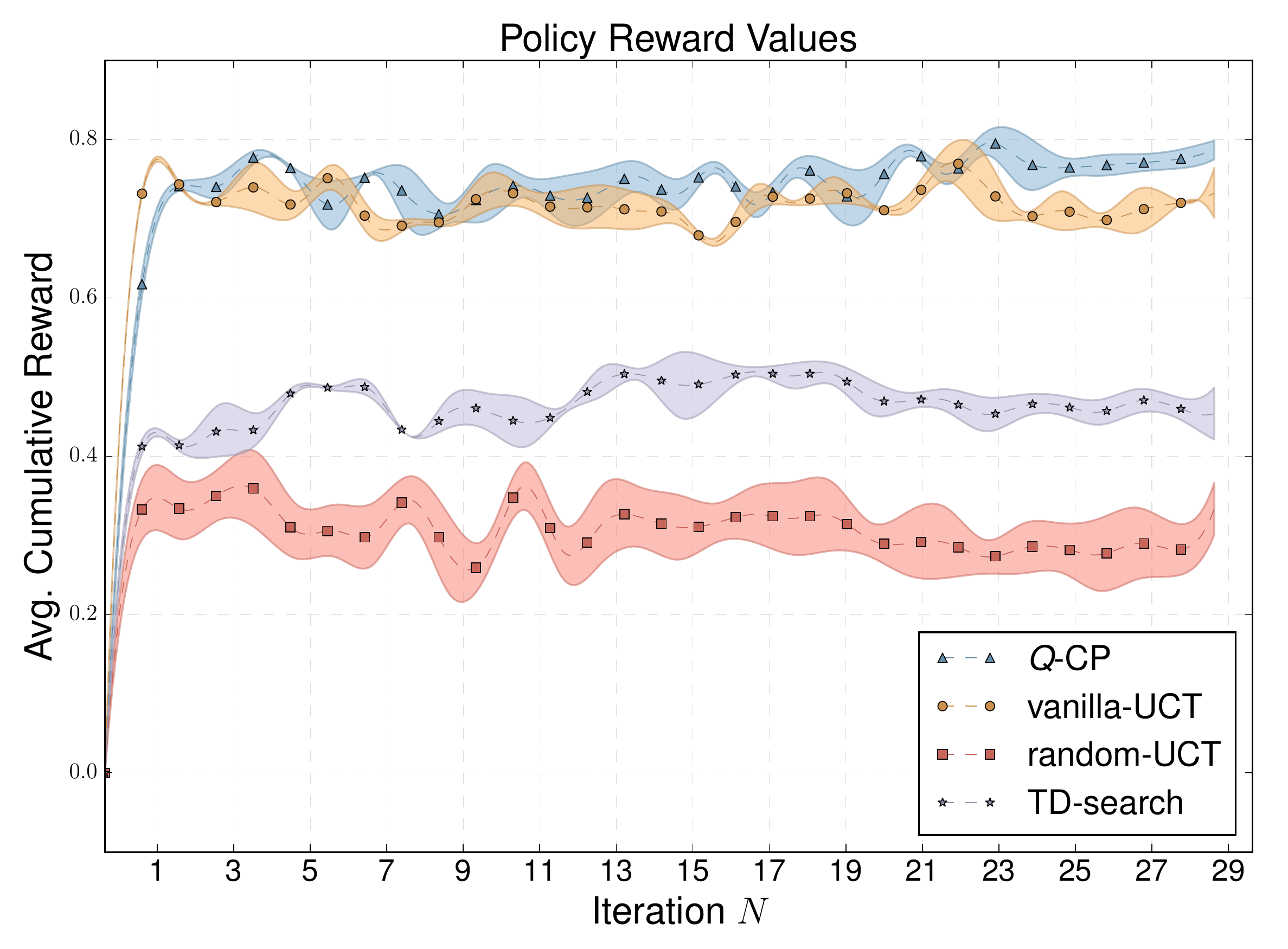}
    \label{fig:door_rewards}
  }
  \subfigure[States] {
    \includegraphics[width=\columnwidth]{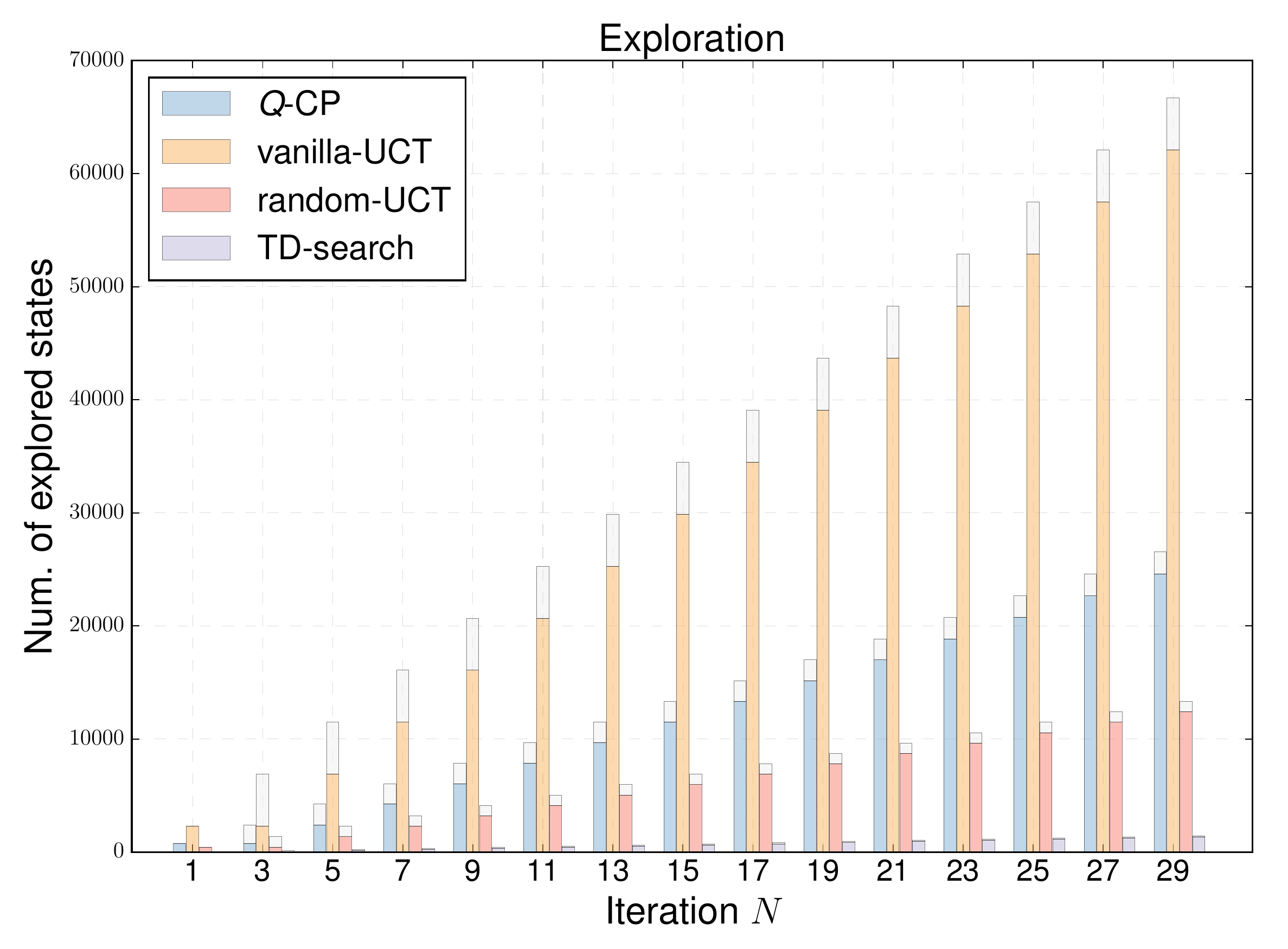}
    \label{fig:door_states}
  }
  \caption{Normalized average reward and number of explored states
    obtained by \qcp{}, TD-search, \textit{random-UCT} and
    \textit{vanilla-UCT} in 29 iterations in the cooperative
    navigation scenario. For each of them, the reward is averaged over
    10 runs.}
  \label{fig:door}
\end{figure}
The last scenario involves timing and precision in completing a
coordination task, where two robots have to traverse a narrow passage,
e.g. a door \figurename~\ref{fig:door_intro}.  Also in this case, the
problem is formalized by discretizing the continuous state space of
the environment. Thus, the state is composed by an 8 features vector
where each robot is represented by its position and the target one, as
$s = \big\langle r_x,~ r_y,~ t_x,~ t_y \big\rangle$. Similarly to the
first scenario, the robots share a set of 5 actions being
$A = \langle$ \texttt{noop, up, down, right, left} $\rangle$.  The
reward function is a weighted sum of 3 components: (1) the first
component in inversely proportional to the distance among robots and
their targets, (2) the second component is proportional to the
distance to obstacles and induces the generated policy to prefer paths
away from them, while (3) the third component is proportional to the
distance between the two robots penilizing states where they are too
close.  \figurename~\ref{fig:door} evaluates \qcp{}, TD-search,
\textit{random-UCT} and \textit{vanilla-UCT} over 29 iterations. The
analysis of the plots is similar to the previous scenarios and
highlights that \qcp{} presents the most suitable trade-off between
performance and computational load, and thus results to be more
practical in robotics applications. In fact, with respect to
\textit{vanilla-UCT}, the number of explored states is still reduced
by the 60\% during last iterations. TD-search, conversely, is not able
to generate a competitive policy by greedily following the
approximated value function (as explained in
Sec~\ref{sec:cooperative_navigation}). This gives a substantial
advantage to \qcp{} in planning complex tasks as the one in this
scenario. It is worth remarking that these plots demonstrate the
ability of \qcp{} to guide the learning routine towards promising
areas of the state search since the first iteration. In fact, \qcp{}
is able to (1) crop out states that do not match the objective of the
task and to (2) generate performing policies even with a reduced
number of samples and iterations.

\section{Conclusion}
\label{sec:conclusion}
In this paper we introduced \qcp, an iterative policy generation
algorithm that uses action values to guide and reduce the exploration
of the state space in different multi-robot scenarios.  Our key
contribution is the combination of a Monte-Carlo tree search algorithm
with action values, which demonstrates a remarkable improvement in the
computational load of the algorithm without loss in the
performance. Such an improvement makes our approach more practical and
suitable in difficult robotics applications, where the lack of
training examples is often a limiting condition.

\qcp{} presents several limitations that give the opportunity to
further improve our solution.  The major problem is the massive use of
simulation calls, that make the algorithm less appealing.  Hence, we
are working on a online version by relying on online and offline
knowledge as in~\cite{Gelly2007}.  Moreover, as the task complexity
increases, the non-linearity of the value function is a structural
problem that affects performance. Thus, we aim at exploiting different
function approximations in order to improve \qcp{} and apply it with
larger state spaces. For example, based on several recent advances in
policy learning \cite{Mnih2015}, we are evaluating the use of neural
networks or deep networks to further improve the performance of our
approach by defining more informed state representations and function
approximators.


\balance
\bibliographystyle{IEEEtranBST/IEEEtran}
\bibliography{IEEEtranBST/IEEEabrv,refs}  

\begin{thebibliography}{10}
\providecommand{\url}[1]{#1}
\csname url@rmstyle\endcsname
\providecommand{\newblock}{\relax}
\providecommand{\bibinfo}[2]{#2}
\providecommand\BIBentrySTDinterwordspacing{\spaceskip=0pt\relax}
\providecommand\BIBentryALTinterwordstretchfactor{4}
\providecommand\BIBentryALTinterwordspacing{\spaceskip=\fontdimen2\font plus
\BIBentryALTinterwordstretchfactor\fontdimen3\font minus
  \fontdimen4\font\relax}
\providecommand\BIBforeignlanguage[2]{{%
\expandafter\ifx\csname l@#1\endcsname\relax
\typeout{** WARNING: IEEEtran.bst: No hyphenation pattern has been}%
\typeout{** loaded for the language `#1'. Using the pattern for}%
\typeout{** the default language instead.}%
\else
\language=\csname l@#1\endcsname
\fi
#2}}

\bibitem{Browne2012}
C.~B. Browne, E.~Powley, D.~Whitehouse, S.~M. Lucas, P.~I. Cowling,
  P.~Rohlfshagen, S.~Tavener, D.~Perez, S.~Samothrakis, and S.~Colton, ``A
  survey of monte carlo tree search methods,'' \emph{IEEE Transactions on
  Computational Intelligence and AI in Games}, vol.~4, no.~1, pp. 1--43, 2012.

\bibitem{Silver2012}
D.~Silver, R.~S. Sutton, and M.~M{\"u}ller, ``Temporal-difference search in
  computer go,'' \emph{Machine learning}, vol.~87, no.~2, pp. 183--219, 2012.

\bibitem{Panait2005}
\BIBentryALTinterwordspacing
L.~Panait and S.~Luke, ``Cooperative multi-agent learning: The state of the
  art,'' \emph{Autonomous Agents and Multi-Agent Systems}, vol.~11, no.~3, pp.
  387--434, Nov. 2005. [Online]. Available:
  \url{http://dx.doi.org/10.1007/s10458-005-2631-2}
\BIBentrySTDinterwordspacing

\bibitem{Kocsis2006}
L.~Kocsis and C.~Szepesv{\'a}ri, ``Bandit based monte-carlo planning,''
  \emph{Machine learning: ECML 2006}, pp. 282--293, 2006.

\bibitem{Agostini2010}
A.~Agostini and E.~Celaya, ``Reinforcement learning with a gaussian mixture
  model,'' in \emph{The 2010 International Joint Conference on Neural
  Networks}, July 2010, pp. 1--8.

\bibitem{Ross2011}
S.~Ross, G.~J. Gordon, and D.~Bagnell, ``A reduction of imitation learning and
  structured prediction to no-regret online learning,'' in \emph{International
  Conference on Artificial Intelligence and Statistics}, 2011, pp. 627--635.

\bibitem{Winands2008}
\BIBentryALTinterwordspacing
M.~H. Winands, Y.~Bj\"{o}rnsson, and J.-T. Saito, ``Monte-carlo tree search
  solver,'' in \emph{Proceedings of the 6th International Conference on
  Computers and Games}, ser. CG '08.\hskip 1em plus 0.5em minus 0.4em\relax
  Berlin, Heidelberg: Springer-Verlag, 2008, pp. 25--36. [Online]. Available:
  \url{http://dx.doi.org/10.1007/978-3-540-87608-3_3}
\BIBentrySTDinterwordspacing

\bibitem{Bowling2000}
M.~Bowling, ``Convergence problems of general-sum multiagent reinforcement
  learning,'' in \emph{In Proceedings of the Seventeenth International
  Conference on Machine Learning}.\hskip 1em plus 0.5em minus 0.4em\relax
  Morgan Kaufmann, 2000, pp. 89--94.

\bibitem{Claus1998}
\BIBentryALTinterwordspacing
C.~Claus and C.~Boutilier, ``The dynamics of reinforcement learning in
  cooperative multiagent systems,'' in \emph{Proceedings of the Fifteenth
  National/Tenth Conference on Artificial Intelligence/Innovative Applications
  of Artificial Intelligence}, ser. AAAI '98/IAAI '98.\hskip 1em plus 0.5em
  minus 0.4em\relax Menlo Park, CA, USA: American Association for Artificial
  Intelligence, 1998, pp. 746--752. [Online]. Available:
  \url{http://dl.acm.org/citation.cfm?id=295240.295800}
\BIBentrySTDinterwordspacing

\bibitem{Kober2013}
J.~Kober, J.~A. Bagnell, and J.~Peters, ``Reinforcement learning in robotics: A
  survey,'' \emph{International Journal of Robotics Research}, July 2013.

\bibitem{Nikolaidis2015}
S.~Nikolaidis, R.~Ramakrishnan, K.~Gu, and J.~Shah, ``Efficient model learning
  from joint-action demonstrations for human-robot collaborative tasks,'' in
  \emph{Proceedings of the Tenth Annual ACM/IEEE International Conference on
  Human-Robot Interaction}, ser. HRI '15.\hskip 1em plus 0.5em minus
  0.4em\relax ACM, 2015, pp. 189--196.

\bibitem{Kim2016}
B.~Kim and J.~Pineau, ``Socially adaptive path planning in human environments
  using inverse reinforcement learning,'' \emph{International Journal of Social
  Robotics}, vol.~8, no.~1, pp. 51--66, 2016.

\bibitem{Sutton1999}
R.~S. Sutton, D.~Precup, and S.~Singh, ``Between mdps and semi-mdps: A
  framework for temporal abstraction in reinforcement learning,''
  \emph{Artificial intelligence}, vol. 112, no. 1-2, pp. 181--211, 1999.

\bibitem{Konidaris2016}
\BIBentryALTinterwordspacing
G.~Konidaris, ``Constructing abstraction hierarchies using a skill-symbol
  loop,'' in \emph{Proceedings of the Twenty-Fifth International Joint
  Conference on Artificial Intelligence}, ser. IJCAI'16.\hskip 1em plus 0.5em
  minus 0.4em\relax AAAI Press, 2016, pp. 1648--1654. [Online]. Available:
  \url{http://dl.acm.org/citation.cfm?id=3060832.3060851}
\BIBentrySTDinterwordspacing

\bibitem{Dietterich2000}
T.~G. Dietterich, ``Hierarchical reinforcement learning with the maxq value
  function decomposition,'' \emph{Journal of Artificial Intelligence Research
  (JAIR)}, vol.~13, pp. 227--303, 2000.

\bibitem{Makar2001}
R.~Makar, S.~Mahadevan, and M.~Ghavamzadeh, ``Hierarchical multi-agent
  reinforcement learning,'' in \emph{Proceedings of the fifth international
  conference on Autonomous agents}.\hskip 1em plus 0.5em minus 0.4em\relax ACM,
  2001, pp. 246--253.

\bibitem{Abtahi2008}
F.~Abtahi and M.~R. Meybodi, ``Solving multi-agent markov decision processes
  using learning automata,'' in \emph{2008 6th International Symposium on
  Intelligent Systems and Informatics}, Sept 2008, pp. 1--6.

\bibitem{HMenell2016}
\BIBentryALTinterwordspacing
D.~Hadfield{-}Menell, A.~D. Dragan, P.~Abbeel, and S.~J. Russell, ``Cooperative
  inverse reinforcement learning,'' \emph{CoRR}, vol. abs/1606.03137, 2016.
  [Online]. Available: \url{http://arxiv.org/abs/1606.03137}
\BIBentrySTDinterwordspacing

\bibitem{Bernstein2014}
\BIBentryALTinterwordspacing
D.~S. Bernstein, C.~Amato, E.~A. Hansen, and S.~Zilberstein, ``Policy iteration
  for decentralized control of markov decision processes,'' \emph{CoRR}, vol.
  abs/1401.3460, 2014. [Online]. Available:
  \url{http://arxiv.org/abs/1401.3460}
\BIBentrySTDinterwordspacing

\bibitem{Proper2009}
S.~Proper and P.~Tadepalli, ``Solving multiagent assignment markov decision
  processes,'' in \emph{Proceedings of The 8th International Conference on
  Autonomous Agents and Multiagent Systems - Volume 1}, ser. AAMAS '09.\hskip
  1em plus 0.5em minus 0.4em\relax Richland, SC: International Foundation for
  Autonomous Agents and Multiagent Systems, 2009, pp. 681--688.

\bibitem{Engel2005}
\BIBentryALTinterwordspacing
Y.~Engel, S.~Mannor, and R.~Meir, ``Reinforcement learning with gaussian
  processes,'' in \emph{Proceedings of the 22Nd International Conference on
  Machine Learning}, ser. ICML '05.\hskip 1em plus 0.5em minus 0.4em\relax New
  York, NY, USA: ACM, 2005, pp. 201--208. [Online]. Available:
  \url{http://doi.acm.org/10.1145/1102351.1102377}
\BIBentrySTDinterwordspacing

\bibitem{Chowdhary2014}
G.~Chowdhary, M.~Liu, R.~Grande, T.~Walsh, J.~How, and L.~Carin, ``Off-policy
  reinforcement learning with gaussian processes,'' \emph{IEEE/CAA Journal of
  Automatica Sinica}, vol.~1, no.~3, pp. 227--238, 2014.

\bibitem{Gelly2011}
\BIBentryALTinterwordspacing
S.~Gelly and D.~Silver, ``Monte-carlo tree search and rapid action value
  estimation in computer go,'' \emph{Artif. Intell.}, vol. 175, no.~11, pp.
  1856--1875, July 2011. [Online]. Available:
  \url{http://dx.doi.org/10.1016/j.artint.2011.03.007}
\BIBentrySTDinterwordspacing

\bibitem{James2017}
S.~James, G.~Konidaris, and B.~Rosman, ``An analysis of monte carlo tree
  search.'' in \emph{AAAI}, 2017, pp. 3576--3582.

\bibitem{Dempster1977}
A.~P. Dempster, N.~M. Laird, and D.~B. Rubin, ``Maximum likelihood from
  incomplete data via the em algorithm,'' \emph{Journal of the Royal
  Statistical Society. Series B (Methodological)}, vol.~39, no.~1, pp. 1--38,
  1977.

\bibitem{MacQueen1967}
J.~MacQueen, ``Some methods for classification and analysis of multivariate
  observations,'' in \emph{Proceedings of the 5th Berkeley Symposium on
  Mathematical Statistics and Probability - Vol. 1}, L.~M. {Le Cam} and
  J.~Neyman, Eds.\hskip 1em plus 0.5em minus 0.4em\relax University of
  California Press, Berkeley, CA, USA, 1967, pp. 281--297.

\bibitem{Schwarz1978}
G.~Schwarz \emph{et~al.}, ``Estimating the dimension of a model,'' \emph{The
  annals of statistics}, vol.~6, no.~2, pp. 461--464, 1978.

\bibitem{Ross2014}
S.~Ross and J.~A. Bagnell, ``Reinforcement and imitation learning via
  interactive no-regret learning,'' \emph{arXiv preprint arXiv:1406.5979},
  2014.

\bibitem{Gelly2007}
S.~Gelly and D.~Silver, ``Combining online and offline knowledge in uct,'' in
  \emph{Proceedings of the 24th international conference on Machine
  learning}.\hskip 1em plus 0.5em minus 0.4em\relax ACM, 2007, pp. 273--280.

\bibitem{Mnih2015}
\BIBentryALTinterwordspacing
V.~Mnih, K.~Kavukcuoglu, D.~Silver, A.~A. Rusu, J.~Veness, M.~G. Bellemare,
  A.~Graves, M.~Riedmiller, A.~K. Fidjeland, G.~Ostrovski, S.~Petersen,
  C.~Beattie, A.~Sadik, I.~Antonoglou, H.~King, D.~Kumaran, D.~Wierstra,
  S.~Legg, and D.~Hassabis, ``Human-level control through deep reinforcement
  learning,'' \emph{Nature}, vol. 518, no. 7540, pp. 529--533, 02 2015.
  [Online]. Available: \url{http://dx.doi.org/10.1038/nature14236}
\BIBentrySTDinterwordspacing

\end{thebibliography}

\end{document}